\documentclass[runningheads]{llncs}
\usepackage{graphicx}

\usepackage{tikz}
\usepackage{comment}
\usepackage{amsmath,amssymb} 
\usepackage{color}
\usepackage{floatrow, caption, tabularx}
\usepackage{color}
\usepackage{float}
\usepackage[export]{adjustbox}
\usepackage{subcaption}
\usepackage{graphics}
\usepackage{makecell}
\usepackage{multirow}
\usepackage{flushend}
\usepackage{xcolor,soul,colortbl}
\usepackage{sidecap}
\usepackage{wrapfig}
\usepackage{booktabs}
\usepackage{enumitem}
\usepackage[linesnumbered,ruled,vlined]{algorithm2e}
\usepackage{algpseudocode}
\SetKwInput{KwInput}{Input} 
\usepackage[pagebackref=true,breaklinks=true,colorlinks,citecolor=blue,bookmarks=false]{hyperref}

\usepackage{tikz}
\usepackage{pgfplots}
\definecolor{mycolor}{RGB}{255,51,76}
\usepackage{pifont}
\newcommand{\xmark}{\ding{55}}%
\newfloatcommand{capbtabbox}{table}[\captop][\FBwidth]

\usepackage[accsupp]{axessibility}  

\newcommand{\std}[1]{\tiny$\pm$ #1}

\makeatletter
\DeclareRobustCommand\onedot{\futurelet\@let@token\@onedot}
\def\@onedot{\ifx\@let@token.\else.\null\fi\xspace}

\def\wrt{w.r.t\onedot}  

\def\bib{BiB\xspace}
\makeatother

\definecolor{Green}{cmyk}{1.,0,1.,0}
\definecolor{NavyBlue}{cmyk}{0.94,0.54,0,0}
\definecolor{Orange}{cmyk}{0,0.61,0.87,0}
\definecolor{RubineRed}{cmyk}{0,1.,0.13,0}
\definecolor{BrickRed}{cmyk}{0,0.89,0.94,0.28}
\definecolor{BlueGreen}{cmyk}{0.85,0,0.33,0}
\definecolor{Brown}{cmyk}{0,0.81,1.,0.60}

\newcommand{\isbib}{\operatorname{is-bib}}
\newcommand{\findbib}{\operatorname{find-bib}}

\newcommand{\paragbf}[1]{\smallskip\noindent\textbf{#1}}

\begin{document}
\pagestyle{headings}
\mainmatter
\def\ECCVSubNumber{1044}  

\title{Active Learning Strategies for Weakly-Supervised Object Detection}

\titlerunning{Active Learning Strategies for Weakly-Supervised Object Detection}
%
\author{
	Huy V. Vo\textsuperscript{1,2}
	\quad
	Oriane Sim{\'e}oni\textsuperscript{2}
	\quad
	Spyros Gidaris\textsuperscript{2}
	\quad
	Andrei Bursuc\textsuperscript{2}
	\quad
	Patrick P{\'e}rez\textsuperscript{2}
	\quad
	Jean Ponce\textsuperscript{1,3}
	\\
	\smallskip
	{\small $^1$Inria and DI/ENS (ENS-PSL, CNRS, Inria) \quad $^2$Valeo.ai \\ $^3$Center for Data Science, New York University}
	\\
	\smallskip
	\small \texttt{\{van-huy.vo, jean.ponce\}@inria.fr} \\ \small \texttt{\{oriane.simeoni,spyros.gidaris,andrei.bursuc,patrick.perez\}@valeo.com}
}
\authorrunning{H. V. Vo et al.}
%
\institute{}
\maketitle
\begin{abstract}
Object detectors trained with weak annotations are affordable alternatives to fully-supervised counterparts. However, there is still a significant performance gap between them. We propose to narrow this gap by fine-tuning a base pre-trained weakly-supervised detector with a few fully-annotated samples automatically selected from the training set using ``box-in-box'' (\bib), a novel active learning strategy designed specifically to address the well-documented failure modes of weakly-supervised detectors. Experiments on the VOC07 and COCO benchmarks show that \bib outperforms other active learning techniques and significantly improves the base weakly-supervised detector's performance with only a few fully-annotated images per class. \bib reaches 97\% of the performance of fully-supervised Fast RCNN with only 10\% of fully-annotated images on VOC07. On COCO, using on average 10 fully-annotated images per class, or equivalently 1\% of the training set, \bib also reduces the performance gap (in AP) between the weakly-supervised detector and the fully-supervised Fast RCNN by over 70\%, showing a good trade-off between performance and data efficiency. Our code is publicly available at \url{https://github.com/huyvvo/BiB}. 

\end{abstract}

\section{Introduction}
Object detectors are critical components of visual perception systems deployed in real-world settings such as robotics or surveillance. Many methods have been developed to build object detectors with high predictive performance \cite{Gidaris2015ICCV_object_detection,girshick2015ICCV_fast_rcnn,girshick2014cvpr_rcnn,He2017ICCV_mask_rcnn,Ren2015NeuRIPS_faster_rcnn} and fast inference~\cite{Redmon2016CVPR_yolo,Redmon2017CVPR_yolo9000}. They typically train a neural network in a fully-supervised manner on large datasets annotated manually with bounding boxes~\cite{pascal-voc-2012,pascal-voc-2007,Lin2014cocodataset}. In practice, the construction of these datasets is a major bottleneck since it involves large, expensive and time-consuming data acquisition, selection and annotation campaigns. To address this challenge, much effort has been put in devising object detection approaches trained with less (or even no) human annotation. This includes semi-supervised~\cite{Jeong2019ConsistencybasedSL,Radosavovic2018DataDT,sohn2020detection,Wang_2018_CVPR}, weakly-supervised~\cite{Bilen2016cvpr_WSDDN,Cinbis2015weakly,Gao2019ICCV_coupled_MIDN_wsod,huang2020comprehensive,ren-cvpr20,Tang2017CVPR_oicr_wsod,zeng2019wsod2}, few-shot~\cite{fan2020few,kang2019few,karlinsky2019repmet,sun2021fsce}, active~\cite{agarwal2020contextual,Brust2019,Choi_2021_ICCV,haussmann2020scalable,sener2018active,settles_active_2009} and unsupervised~\cite{Cho2015ObjectDiscovery,LOST,Sivic05ObjectDiscovery,TangLewis2008uod,Vo2019UnsupOptim,Vo21LOD} learning frameworks for object detection.

\begin{figure}[htb]
    \centering
    \includegraphics[width=\textwidth]{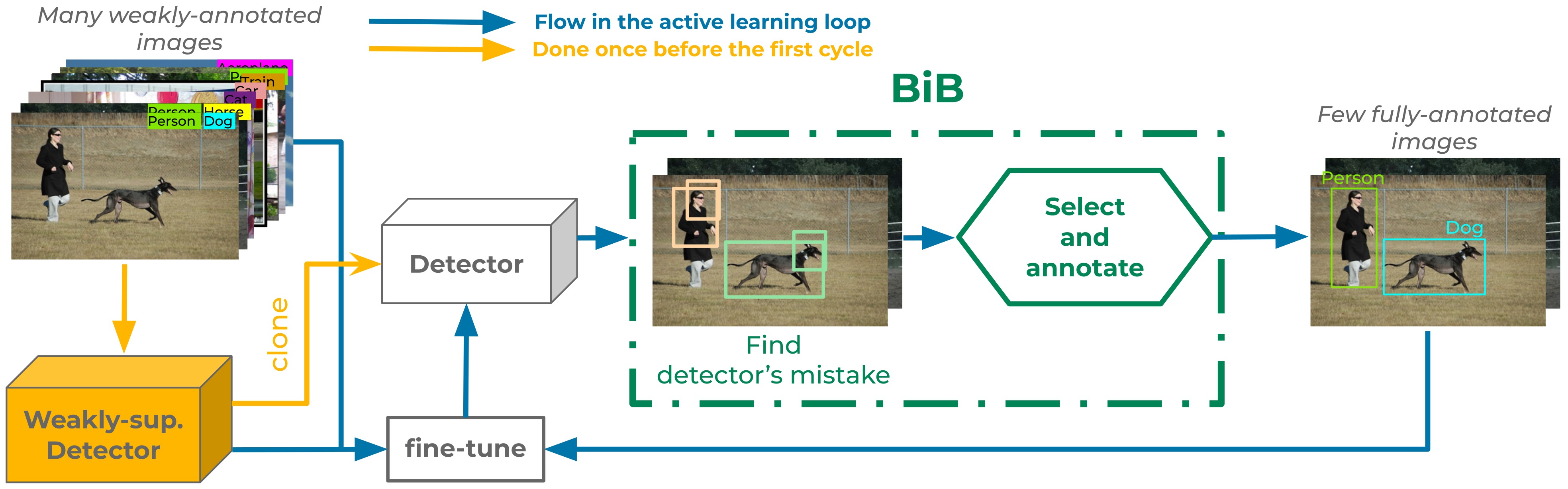}
    \caption{\small Overview of our approach. A base object detector is first trained only with image-level tags, then fine-tuned in successive stages using few \emph{well-selected} images that are fully annotated. For their selection, we propose \emph{``box-in-box''} (\bib), an acquisition function designed to discover recurring failure cases of the weakly-supervised detector, e.g., failure to localize whole objects or to separate distinct instances of the same class. }
    \label{fig:intro}
\end{figure}

Weakly-supervised object detection (WSOD) typically only uses image-level category labels during training~\cite{Bilen2016cvpr_WSDDN,ren-cvpr20,Tang2017CVPR_oicr_wsod}. This type of annotation is much cheaper than bounding boxes and, in some cases, it can be even obtained automatically, e.g., leveraging tags on online photos, photo captions in media or time-stamped movie scripts. WSOD is thus an affordable alternative to fully-supervised object detection in terms of annotation cost. However, weakly-supervised detectors often struggle to correctly localize the full extent of objects~\cite{ren-cvpr20,Tang2017CVPR_oicr_wsod}.
Several recent works~\cite{Biffi2020ManyshotFL,Pan2019box_correction} show that a good trade-off between performance and annotation cost can be achieved by annotating bounding boxes in a small set of randomly selected training images and by training the detector with a mix of weak and full supervision. However, there are better alternatives to random selection. Active learning (AL) methods \cite{Choi_2021_ICCV,MIAOD2021} offer means to \emph{select} images that should be the most helpful for the training of an object detector model, given some criterion.

In this work, we propose to combine both worlds, by augmenting the weakly-supervised framework with an active learning scheme. Our active learning strategy specifically targets the known failure modes of weakly-supervised detectors. We show that it can be used to significantly narrow the gap between weakly-supervised detectors and expensive fully-supervised ones with a few fully-annotated images per class. We start with a weakly-annotated dataset, e.g., a set of images and their class labels, with which we train a weakly-supervised detector. We apply our new active learning strategy that we call \emph{box-in-box} (\bib) to iteratively select from the dataset a few images to be fully annotated. New full annotations are added to the training set and used to fine-tune the detector. Given the fine-tuned detector, we select another batch of images to be fully annotated. This process is repeated several times to improve the detector (\textcolor{red}{Figure}~\ref{fig:intro}). Previous works have attempted to combine weak supervision with active learning, but they all start with an initial set of hundreds to thousands of fully-annotated images. As shown in \textcolor{red}{Section}~\ref{sec:exp}, our approach only requires a small number of fully-annotated images ($50-250$ on VOC07~\cite{pascal-voc-2007} and $160-800$ on COCO~\cite{Lin2014cocodataset}) to significantly improve the performance of weakly-supervised detectors. 
Our main contributions are: 
(1) We propose a new approach to improve weakly-supervised object detectors, by using a few fully-annotated images, carefully selected with the help of active learning. Contrary to typical active learning approaches, we initiate the learning process without any fully-annotated data;
(2) We introduce \bib, an active selection strategy that is tailored to address the limitations of weakly-supervised detectors;
(3) We validate our proposed approach with extensive experiments on VOC07 and COCO datasets. We show that \bib outperforms other active strategies on both datasets, and reduce significantly the performance gap between weakly- and fully-supervised object detectors.

\section{Related Work}

\paragbf{Weakly-supervised object detection} is a data-efficient alternative to fully-supervised object detection which only requires image-level labels (object categories) for training a detector. It is typically formulated as a multiple instance learning problem~\cite{dietterich1997mil}
where images are bags and region proposals~\cite{uijlings2013selective,zitnick2014edge} are instances. The model is trained to classify images using scores aggregated from their regions, through which it also learns to distinguish {\em object} from {\em non-object} regions. 
Since training involves solving a non-convex optimization problem, adapted initialization and regularization techniques~\cite{Cinbis2015weakly,Deselaers2010localize,Kumar2010selfpace,song2014learning,Song2014weakly} are necessary for good performance. Bilen {\em et al.}~\cite{Bilen2016cvpr_WSDDN} proposes WSDDN, a CNN-based model for WSOD which is improved in subsequent works~\cite{Diba2017CVPR_cascaded_net_wsod,Jie2017deepselftaught,ren-cvpr20,Tang2018TPAMI_pcl_wsod,Tang2017CVPR_oicr_wsod}. Tang {\em et al.}~\cite{Tang2017CVPR_oicr_wsod} proposes OICR which refines WSDDN's output with parallel detector heads in a self-training fashion. Trained with only image-level labels, weakly-supervised object detectors are often confused between object parts and objects, or between objects and groups of objects~\cite{ren-cvpr20}. Although mitigating efforts with better pseudo labels~\cite{ren-cvpr20,Tang2018TPAMI_pcl_wsod}, better representation~\cite{huang2020comprehensive,ren-cvpr20} or better optimization~\cite{Arun2019CVPR_dissim_coef_wsod,Wan2019CVPR_CMIL_wsod} achieve certain successes, such confusion issues of weakly-supervised detectors remain due to the lack of a formal definition of objects and their performance is still far behind that of fully-supervised counterparts. In this work, we show that fine-tuning weakly-supervised detectors with strong annotation on \emph{a few carefully selected} images can alleviate these limitations and significantly narrow the gap between weakly- and fully-supervised object detectors.  

\paragbf{Semi-supervised object detection} methods exploits a mix of some fully-annotated and many unlabelled-data. Two dominant strategies arise among these methods: consistency-based~\cite{Jeong2019ConsistencybasedSL,Tang2021ProposalLF} and pseudo-labeling~\cite{Li2020ImprovingOD,Radosavovic2018DataDT,sohn2020detection,Wang_2018_CVPR,xu2021end,Zoph2020RethinkingPA}. The latter can be further extended with strategies inspired by active learning~\cite{Li2020ImprovingOD,Wang_2018_CVPR} for selecting boxes to be annotated by humans. 

\paragbf{Combining weakly- and semi-supervised object detection.} These approaches seek a better trade-off between performance and annotation cost than individual strategies. All images from the training set have weak labels and a subset is also annotated with bounding boxes. This setup enables the exploration of the utility of additional types of weak labels, e.g., points~\cite{Chen2021PointsAQ,ren-eccv2020} or scribbles~\cite{ren-eccv2020}. Others leverage fully-annotated images to train detectors that can correct wrong predictions of weakly-supervised detectors~\cite{Pan2019box_correction} or compute more reliable pseudo-boxes~\cite{Biffi2020ManyshotFL}. 
Similarly to \cite{Biffi2020ManyshotFL,Pan2019box_correction}, we train a detector with only a few annotated images, but different from them, we focus on how to best select the images to annotate towards maximizing the performance of the detector.  

\paragbf{Active learning for object detection} aims at carefully \emph{selecting} images to be fully annotated by humans, in order to minimize human annotation efforts. Most methods exploit \emph{data diversity}~\cite{Geifman2017DeepAL,sener2018active} or \emph{model uncertainty}~\cite{Brust2019,Choi_2021_ICCV} to identify such images. These strategies, originally designed for generic classification tasks~\cite{settles_active_2009}, have been recently derived and adapted to  object detection~\cite{Choi_2021_ICCV,MIAOD2021}, a complex task involving both classification (object class) and regression (bounding box location). Data diversity can be ensured by selecting data samples using image features and applying k-means~\cite{Zhdanov2019DiverseMA}, k-means++ initialization~\cite{haussmann2020scalable} or identifying a core-set -- a \emph{representative} subset of a dataset~\cite{agarwal2020contextual,Geifman2017DeepAL,sener2018active}. Model uncertainty for AL can be computed from image-level scores aggregated from class predictions over boxes~\cite{Brust2019,haussmann2020scalable,Pardo2021BAODBO}, comparing predictions of the same image from different corrupted versions of it~\cite{elezi21,Kao2018LocalizationAwareAL,Gao2020Consistency} or from different steps of model training \cite{huang2021semi,Soumya2018}, voting over predictions from an ensemble of networks~\cite{beluch2018power,chitta2019large,haussmann2020scalable}, Bayesian Neural Networks~\cite{gal2017deep,haussmann2020scalable} or single forward networks mimicking an ensemble~\cite{Choi_2021_ICCV,MIAOD2021}. Multiple other strategies have been proposed for selecting informative, difficult or confusing samples to annotate by: learning to discriminate between labeled and unlabeled data~\cite{Ebrahimi2020MinimaxAL,Ebrahimi2020VariationalAA,Gissin2019DiscriminativeAL,Zhang2020StateRelabelingAA}, learning to predict the detection loss~\cite{L4AL19}, the gradients~\cite{Ash2020Deep} or the influence of data on gradient~\cite{liu2021influence}.  
In contrast to classical active learning methods in which the initial model is trained in a fully-supervised fashion using a randomly sampled initial set of images, our initial model is only trained with weakly-annotated data. This is a challenging problem, but often encountered in practice when new collections of data arrive only with weak annotations and significant effort is required to select which images to annotate manually prior to active learning. 

\paragbf{Combining weak supervision and active learning.} 
Closer to us, \cite{Desai2019AnAS,Fang_Xu_Liu_Parisot_Li_2020,Pardo2021BAODBO} investigate how weakly-supervised learning and active learning can be conducted together in the context of object detection. Desai et al.~\cite{Desai2019AnAS} propose to use clicks in the center of the object as weak labels which include localization information and are stronger than image-level tags. Pardo et al.~\cite{Pardo2021BAODBO} also mix strong supervision, tags and pseudo-labels in an active learning scenario. Both \cite{Desai2019AnAS,Pardo2021BAODBO} rely on Faster R-CNN~\cite{Ren2015NeuRIPS_faster_rcnn} and \cite{Fang_Xu_Liu_Parisot_Li_2020} on a FPN~\cite{Lin2017FeaturePN} -- detectors that are hard to train only with weak labels. All start with $10\%$ of the dataset fully labeled, which is more than the total amount of fully annotated data we consider in this work. 

\section{Proposed Approach}

\subsection{Problem Statement}
We assume that we are given $n$ images $\mathcal{I}=\{\mathbf{I}_i\}_{i \in \{1\ldots n\}}$ annotated with labels $\mathcal{Q}=\{\mathbf{q}_i\}_{i \in \{1\ldots n\}}$. Here $\mathbf{q}_i \in \{0,1\}^{C}$ is the class label of image $\mathbf{I}_i$, with $C$ being the number of classes in the dataset. Let $M^0$ be a weakly-supervised object detector trained using only $\mathcal{Q}$. The goal of our work is to iteratively select a {\em very small set of images} to fully annotate with bounding boxes and fine-tune $M^0$ on the same images with both weak and full annotation so as to maximize its performance. To that end, we propose a novel \emph{active learning} method properly adapted to the aforementioned problem setting.

\subsection{Active Learning for Weakly-Supervised Object Detection}

As typical in active learning, our approach consists of several cycles in which an acquisition function first uses the available detector to select images that are subsequently annotated by a human with bounding boxes. The model is then updated with this additional data. With the updated detector, a new cycle of acquisition is performed (see \textcolor{red}{Algorithm}~\ref{algo:w+al}). 

\begin{algorithm}[t]
\SetAlgoLined
\SetKwFunction{Label}{label}
\SetKwFunction{Train}{train}
\SetKwFunction{Fn}{fine-tune}
\SetKwFunction{Select}{select}
\KwInput{Set $\mathcal{I}$ of weakly-labelled images, set $\mathcal{Q}$ of weak annotations, number of cycles $T$, budget per cycle $B$.}
\KwResult{Detector $M^T$, bounding box annotations $\mathcal{G}^T$.} 

$M^0 \gets \Train(\mathcal{I}, \mathcal{Q})$ \Comment{weakly-supervised pre-training} \\
\For{$t = 1$ \text{to} $T$} 
    {
    $A^t \gets \Select$($W^{t-1}$, $M^{t-1},\mathcal{I}, \mathcal{Q}, B$) \Comment{select a batch $A^t$} of $B$ images\\
    
    $\mathcal{G}^{t} \gets \mathcal{G}^{t-1} \cup \Label(\mathcal{I}, A^t)$ \Comment{annotate new selection} \\
    $S^t \gets S^{t-1} \cup A^{t}$, ~$W^t \gets W^{t-1} \setminus A^{t}$ \Comment{update the sets} \\ 
    
    $M^t \gets \Fn (\mathcal{I}, \mathcal{Q}, \mathcal{G}^t, M^0)$ \Comment{fine-tune the model} \\
}
\caption{WSOD with Active Learning.}
\label{algo:w+al}
\end{algorithm}
Let $W^t \subset \{1,\ldots,n\}$ be the set of indices of images with class labels only, and $S^t \subset \{1,\ldots,n\}$ the set with bounding-box annotations at the $t$-th active learning cycle.
The active learning process starts with $W^0 = \{1,\ldots,n\}$ and $S^0 = \varnothing$. Then, at each cycle $t > 0$, the acquisition function selects from $ W^{t-1}$ a set $A^t$ of $B$ images to be annotated with bounding boxes, with $B$ the fixed annotation budget per cycle. By definition, we have that $A^t \subset W^{t-1}$ and $|A^t| = B$.
For the selection, the acquisition function exploits the detector $M^{t-1}$ obtained at the end of the previous cycle. After selecting $A^t$, the sets of fully and weakly-annotated images are updated with $S^{t} = S^{t-1} \cup A^t$ and $W^{t} = W^{t-1} \setminus A^t$ respectively. We define as $\mathcal{G}^t = \{\mathbf{G}_{i}\}_{i \in S^t}$ the bounding-box annotations for images with indices in $S^t$. Finally, at the end of cycle $t$, we fine-tune $M^0$ on the entire dataset, using the bounding box annotations for images with indices in $S^t$ and the original image-level annotations for others. 

\subsection{\bib: An Active Learning Strategy}
\label{sec:bib_al}

\begin{figure}[t]
    \centering
    \includegraphics[width=\linewidth,height=0.42\linewidth]{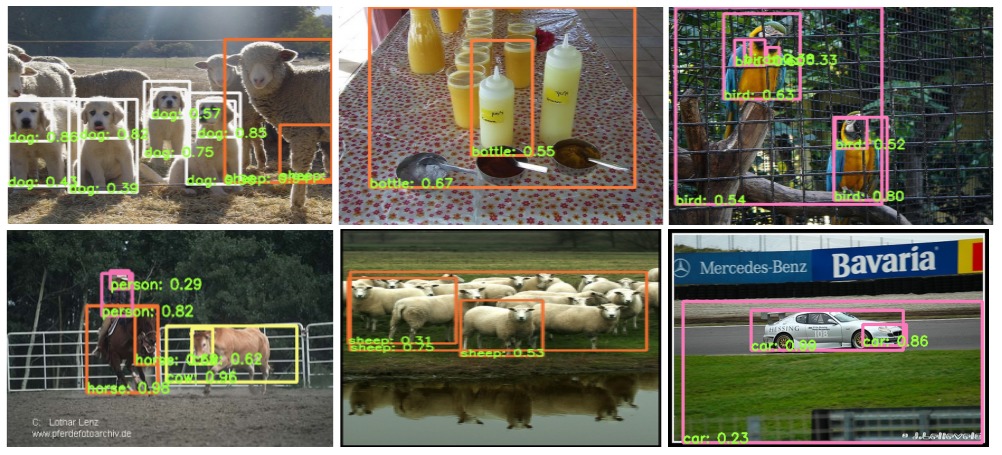}
    \caption{\small Example of box-in-box (BiB) pairs among the predictions of the weakly-supervised object detector. The existence of such pairs is an indicator of the detector's failure on those images. Best viewed in color.}
    \label{fig:box_in_box_pairs}
\end{figure}
With a very small annotation budget, we aim at selecting the ``best" training examples to ``fix" the mistakes of the base weakly-supervised detector. We propose \emph{\bib}, an acquisition strategy tailored for this purpose. It first \emph{discovers (likely) detection mistakes} of the weakly-supervised detector, and then selects \emph{diverse detection mistakes}. Our selection strategy is summarized in \textcolor{red}{Algorithm}~\ref{algo:BiB}.

\paragbf{Discovering \bib patterns.} Weakly-supervised detectors often fail to accurately localize the full extent of the objects in an image, and tend to focus instead on the most \emph{discriminative parts} of an object or to group together multiple object instances~\cite{ren-cvpr20}. Several examples of these errors are shown in \autoref{fig:box_in_box_pairs}.
In the first column, a predicted box focuses on the most discriminative part of an object while a bigger one encompasses a much larger portion of the same object. Another recurring mistake is when two or more distinct objects are grouped together in a box, but some correct individual predictions are also provided for the same class (second column). The two kinds of mistakes can also be found in the same image (third column). We name ``box-in-box" (\bib) such detection patterns where two boxes are predicted for a same object class, a small one being ``contained'' (within some tolerance, see below) in a larger one. We take BiB pairs as an indicator of model's confusion on images.

More formally, let $\mathbf{D}_i$ be the set of boxes detected in image $\mathbf{I}_i$ and let $d_A$ and $d_B$ be two of them.  We consider that $(d_A, d_B)$ is a \bib pair, which we denote with $\isbib(d_A, d_B) = \texttt{True}$, when:
(i) $d_A$ and $d_B$ are predicted for the same class, (ii) $d_B$ is at least $\mu$ times larger than $d_A$ (i.e., $\frac{\mathrm{Area}(d_B)}{\mathrm{Area}(d_A)} \geq  \mu$), 
and (iii) the intersection of $d_B$ and $d_A$ over the area of $d_A$ is at least $\delta$ (i.e., $\frac{\mathrm{Intersection}(d_A,d_B)}{\mathrm{Area}(d_A)} \geq \delta$). Hence, the set $P_i = \{p_{i,j}\}_{j=1}^{|P_i|}$ of \bib pairs is found in image $\mathbf{I}_i$ by the procedure
\begin{equation}
\label{eq:findbib}
 \findbib(\mathbf{D}_i) = \{(d_A,d_B) \in \mathbf{D}_i\times \mathbf{D}_i | \isbib(d_A, d_B) \}   \text{.}
\end{equation}
We observe that in such a \bib pair, it is likely that at least one of the boxes is a detection mistake. We thus propose to select images to be fully annotated among those containing \bib pairs. 

\begin{algorithm}[t]
\SetAlgoLined
\SetKwFunction{Pairs}{get-bib-pairs}
\SetKwFunction{Most}{most-pairs}
\SetKwFunction{Detect}{Detect}
\SetKwFunction{GetImageIndex}{get-imid}
\SetKwFunction{Prob}{Prob}
\SetKwFunction{Feature}{extract-features}
\SetKwFunction{Random}{random-sampling}
\KwResult{Set $A^t$ of selected images.} 
\KwInput{Budget $B$, model $M^{t-1}$, image set $\mathcal{I}$, 
        index set $W^{t-1}$ of weakly-annotated images, 
        set $\hat{\mathcal{P}}$ of already selected \bib pairs (if empty, see text for initialization)}

\For{$i \in W^{t-1}$}{
    $\mathbf{D}_i \gets \Detect(\mathbf{I}_i | M^{t-1})$  \Comment{Predict boxes} \\
    $P_i \gets \{p_{i,j}\}_{j=1}^{|P_i|} = \findbib(\mathbf{D}_i)$ \Comment{Discover \bib patterns}
}

\# Select diverse detection mistakes \\
$A^t \gets \varnothing$\\
\While{$|A^t| < B$}{
        \For{$p \in \cup_{i \in W^{t-1} \setminus A^t} P_i$}{
            $w_p \gets \min_{\tilde{p} \in \hat{\mathcal{P}}} \|F(p) - F(\tilde{p})\|$ \Comment{Comp. dist. to selected pairs} \\
        }
        $p^{\ast} \sim \text{Prob}(\{{w}_p\}_p)$ \Comment{Randomly select a pair}\\
        $i^{\ast} \gets \GetImageIndex(p)$ \Comment{Get index of the image containing p} \\
        $\hat{\mathcal{P}} \gets \hat{\mathcal{P}} \cup P_{i^{\ast}}$, $A^t \gets A^t \cup \{i^{\ast}\}$ \Comment{Updates}
}
\caption{\bib acquisition strategy.}
\label{algo:BiB}
\end{algorithm}

\paragbf{Selecting diverse detection mistakes.} Given the set of all \bib pairs over $\mathcal{I}$, the acquisition function considers the \emph{diversity} of the pairs in order to select images. In particular, we follow \textit{k-means++ initialization}~\cite{arthurKmeans} -- initially developed to provide a good initialization to k-means clustering by iteratively selecting as centroids data points that lie further away from the current set of selected ones. This initialization has previously been applied onto image features in the context of active learning for object detection \cite{haussmann2020scalable} or on model's gradients for active learning applied to image classification \cite{Ash2020Deep}. Here we focus and apply the algorithm to pairs of detected boxes.

We denote with $\hat{\mathcal{P}}$ the set of \bib pairs from the already selected images. For each pair $p$ not in $\hat{\mathcal{P}}$, we compute the minimum distance $w_p$ to the pairs in $\hat{\mathcal{P}}$: $w_p \gets \min_{\tilde{p} \in \hat{\mathcal{P}}} \|F(p) - F(\tilde{p})\|$, where $F(p)$ is the feature vector of $p$, 
i.e., the concatenation of the region features corresponding to the two boxes of $p$ each extracted using the model $M^{t-1}$.
We then randomly pick a new pair $p^{\ast}$, using a weighted probability distribution where a pair $p$ is chosen with probability proportional to $w_p$. We finally select the image $\mathbf{I}_{i^{\ast}}$ that contains $p^{\ast}$, add its index $i^{\ast}$ to $A^t$ and its BiB pairs to $\hat{\mathcal{P}}$. Note that at the beginning of the selection process in each cycle, $\hat{\mathcal{P}}$ contains the pairs of images selected in the previous cycles and is empty when the first cycle begins. In the latter case, we start by selecting the image $\mathbf{I}_{i^{\ast}}$ that has the greatest number of pairs $|P_{i^{\ast}}|$\footnote{In case of a draw, an image is randomly selected.} and add the pairs in $P_{i^{\ast}}$ to $\hat{\mathcal{P}}$ before starting the selection process above.

With this design, \bib selects a diverse set of images that are representative of the dataset while addressing the known mistakes of the weakly-supervised detector. We show some examples selected by \bib and demonstrate its effectiveness in boosting the performance of the weakly-supervised detector in \textcolor{red}{Section}~\ref{sec:exp}. 

\subsection{Training Detectors with both Weak and Strong Supervision}
\label{sec:weakly}
We provide below details about the step of model fine-tuning performed at each cycle. For clarity, we drop the image index $i$ and the cycle index $t$ in this section.

\paragbf{Training with weak annotations.} We adopt the state-of-the-art weakly-supervised method MIST~\cite{ren-cvpr20} as our base detector. MIST follows~\cite{Tang2017CVPR_oicr_wsod} which adapts the detection paradigm of Fast R-CNN~\cite{girshick2015ICCV_fast_rcnn} to weak annotations. It leverages pre-computed region proposals extracted from unsupervised proposal algorithms, such as Selective Search~\cite{uijlings2013selective} and EdgeBoxes~\cite{zitnick2014edge}. In particular, given image $\mathbf{I}$ which has only weak labels $q$ (class labels) and its set of region proposals $\mathcal{R}$, simply called regions, the detection network extracts the image features with a CNN backbone and compute for each region a feature vector using a region-wise pooling~\cite{girshick2015ICCV_fast_rcnn}. Then, the network head(s) on top of the CNN backbone process the extracted region features in order to predict for each of them the object class and modified box coordinates. To build a detector that can be effectively trained using only image-wise labels, MIST has two learning stages, \emph{coarse detection with multiple instance learning} and \emph{detection refinement with pseudo-boxes}, each implemented with different heads but trained simultaneously in an online fashion~\cite{Tang2017CVPR_oicr_wsod}. 

The \emph{Multiple Instance Learner} (MIL) head is trained to minimize the multi-label classification loss $\mathcal{L}^{\text{MIL}}$ using weak labels through which it produces classification scores for all regions in $\mathcal{R}$. MIST selects from them the regions with the highest scores (with non-maximum suppression) as coarse predictions, which we denote with $\mathbf{D}^{(0)}$. Then, such predictions are iteratively refined using $K$ consecutive \emph{refinement heads}. Each refinement head $k \in \{1\ldots K\}$ predicts for all regions in $\mathcal{R}$ their classification scores for the $C+1$ classes ($C$ object classes plus $1$ background class) and box coordinates per object class. The refinement head $k$ is trained by minimizing: 
\begin{equation}
    \label{eq:refinement}
    \mathcal{L}^{(k)}_w(\mathbf{I}, \mathcal{R}, \mathbf{D}^{(k-1)}) = \mathcal{L}_{\text{cls}}^{(k)}(\mathbf{I}, \mathcal{R}, \mathbf{D}^{(k-1)})+ \mathcal{L}_{\text{reg}}^{(k)}(\mathbf{I}, \mathcal{R}, \mathbf{D}^{(k-1)}),
\end{equation}
which combines an adapted instance classification loss, $\mathcal{L}_{\text{cls}}^{(k)}$, and the box regression loss $\mathcal{L}_{\text{reg}}^{(k)}$ of Fast R-CNN~\cite{girshick2015ICCV_fast_rcnn}, 
using as targets the pseudo-boxes $\mathbf{D}^{(k-1)}$ generated by MIST from the region scores of the previous head. The final loss for image $\mathbf{I}$ is:
\begin{equation}
    \mathcal{L}_w = \mathcal{L}^{\text{MIL}}(\mathbf{I}, \mathcal{R}, \mathbf{q})  + \sum_{k=1}^K \mathcal{L}^{(k)}_w(\mathbf{I}, \mathcal{R}, \mathbf{D}^{(k-1)}).
\end{equation}
For more details about MIST, please refer to the appendix and~\cite{ren-cvpr20}. 

\paragbf{Adding strong annotations.}
\label{sec:strong}
In our proposed approach, we obtain ground-truth bounding boxes for \emph{very few} images in the set $\mathcal{I}$. In order to integrate such strong annotations to the weakly-supervised framework, we simply replace the pseudo-annotations in \textcolor{red}{Equation}~\ref{eq:refinement} with box annotations $\mathbf{G}$, now supposed available for image $\mathbf{I}$. The resulting loss for the refinement head $k$ reads 
$\mathcal{L}^{(k)}_s(\mathbf{I}, \mathcal{R}, \mathbf{G}) = \mathcal{L}_{\text{cls}}^{(k)}(\mathbf{I}, \mathcal{R}, \mathbf{G}) + \mathcal{L}_{\text{reg}}^{(k)}(\mathbf{I}, \mathcal{R}, \mathbf{G})$,
and the final loss on image $\mathbf{I}$ in this case is 
$\mathcal{L}_s = \mathcal{L}^{\text{MIL}}(\mathbf{I}, \mathcal{R}, \mathbf{q})  + \sum_{k=1}^K \mathcal{L}^{(k)}_s(\mathbf{I}, \mathcal{R}, \mathbf{G})$.
As such, during the fine-tuning of the detector $M^0$, we use $\mathcal{L}_w$ to train on images for which only class labels are available and $\mathcal{L}_s$ when images are provided with
bounding-boxes. 

\paragbf{Difficulty-aware proposal sampling.}
\label{sec:dificulty-aware}
In this framework, we use thousands of pre-computed proposals in $\mathcal{R}$ for each image. This is necessary when no box annotations are provided. However, when ground-truth boxes are available, better training can be achieved by sampling a smaller number of proposals \cite{girshick2015ICCV_fast_rcnn,ren-eccv2020}. In particular, we select a subset of $512$ proposals with $25\%$ of \emph{positive} boxes, i.e., those whose IoU with one of the ground-truth boxes exceeds $0.5$, and $75\%$ of \emph{negative} boxes, i.e., those whose IoU $\leq 0.3$ with all ground-truth boxes. However, we have noticed that negatives are over-sampled from the background or often appear uninformative. We propose to improve negative proposal sampling by using the network predictions to select those classified as objects. We perform a first forward pass and average classification scores obtained over the $K$ refinement heads; we then apply row-wise softmax and select proposals with the highest class scores, excluding background. We show in our experiments that this sampling method allows better training and yields better performance. 

\section{Experimental Results}
\label{sec:exp}
In this section, we first introduce the general setup of our experiments. We then present an ablation study of different components of \bib before comparing \bib to different existing active learning strategies. Finally, we demonstrate the effectiveness of our proposed pipeline compared to the state of the art.

\subsection{Experimental Setup}

\paragbf{Datasets and evaluation.} We evaluate our method on two well-known object detection datasets, Pascal VOC2007 \cite{pascal-voc-2007} (noted VOC07) and COCO2014 \cite{Lin2014cocodataset} (COCO). Following previous works~\cite{Biffi2020ManyshotFL,ren-cvpr20}, we use the {\em trainval} split of VOC07 for training and the {\em test} split for evaluation, respectively containing 5011 and 4952 images. On COCO, we train detectors with the {\em train} split (82783 images) and evaluate on the {\em validation} split (40504 images) following \cite{Biffi2020ManyshotFL}. We use the average precision metrics AP50 and AP, computed respectively with an IoU threshold of 0.5 and with thresholds in $[0.5:0.95]$. 
We report results corresponding to $N$-shot experiments -- where $N\times C$ images are selected -- and $N\%$ experiments, where about $N$ percents of the training set are selected to be fully-annotated.

\paragbf{Architecture.} Though \bib can be applied on any weakly-supervised detector, we use MIST~\cite{ren-cvpr20} as our base weakly-supervised detector for it has public code and has been shown to be a strong baseline. We modify MIST to account for images containing bounding box annotations during training as detailed in \textcolor{red}{Section}~\ref{sec:strong}. The Concrete Drop Block (CDB)~\cite{ren-cvpr20} technique is used in MIST in experiments on VOC07 but dropped in COCO experiments to save computational cost. We use our difficulty-aware proposal sampling in all experiments unless stated otherwise. We train with a batch size of $8$ during training and a learning rate of $4\mathrm{e}{-4}$ for MIST and $4\mathrm{e}{-6}$ for CDB when the latter is used. During training, images are drawn from the sets of images with weak and strong annotation uniformly at random such that the numbers of weakly- and fully-annotated images considered are asymptotically equal.

\paragbf{Active learning setup.} We emulate an active learning setup by ignoring available bounding box annotations of images considered weakly annotated in our experiments. On both dataset, we run MIST~\cite{ren-cvpr20} three times to account for the training's instability and obtain three base weakly-supervised detectors. We fine-tune each base weakly-supervised detector twice on VOC07 and once on COCO, giving respectively 6 and 3 repetitions. We always report averaged results, and in some cases also their standard deviation. Detailed results for all experiments are provided in supplemental material. The number of fine-tuning iterations is scaled linearly with the number of strong images in the experiment. Concretely, the base weakly-supervised detector is fine-tuned over $300$ iterations for every $50$ strong images in VOC07 and $1200$ iterations for every $160$ images on COCO.

\paragbf{Active learning baselines.} We compare \bib to existing active learning strategies. We first compare our method to random selections, either uniform sampling (\emph{u-random}) or balanced per class sampling (\emph{b-random}). We compare to uncertainty-based selection and aggregate box entropy scores per image using sum or max pooling, noted {\em entropy-sum} and {\em entropy-max} respectively. Finally, we leverage weak detection losses to select high impact images ({\em loss}). We report here results obtained with the detection loss of the last refinement branch in MIST, which we find outperforms others losses; a detailed comparison can be found in supplemental material along with a complete description of other methods. 
We also use the greedy version of the \emph{core-set} selection method ~\cite{sener2018active}; and a weighted version that weights distances in core-set with uncertainty scores (entropy-max)~\cite{haussmann2020scalable}, named \emph{coreset-ent}.  For our \bib, we set $\mu = 3$ and $\delta= 0.8$, and provide a study on their influence in the supplemental material.

\subsection{Ablation Studies}
We perform in \autoref{table:ablation_study} an ablation study to understand the relative importance of the difficulty-aware proposal sampling (\emph{DifS}), the selection based on k-means++ initialization and the use of box-in-box pairs in our method. The second row corresponds to \emph{u-random}. We apply the diversity selection (e.g., following k-means++ initialization) on image-level features, predictions, and \bib pairs. The experiments are conducted on VOC07 and for each variant of our method, we perform five active learning cycles with a budget of $50$ images per cycle. It appears that \emph{DifS} significantly improves results over both random and \bib, confirming that targeting the model's most confusing regions is helpful. K-means++ initialization does not help when applied on image-level features but yields significant performance boosts over random when combined with region-level features. Finally, the use of \bib pairs shows consistent improvements over {\em region}, confirming our choices in \bib's design.

\begin{table}[t]
\resizebox{0.8\textwidth}{!}{
    \caption{\small 
    Ablation study. Results in AP50 on VOC07 with $5$ cycles and a budget $B = 50$. We provide averages and standard deviation results over 6 repetitions. \emph{DifS} stands for the difficulty-aware region sampling module. Images are selected by applying k-means++ init. ({\em K selection}) on image-level features ({\em im.}), confident predictions' features ({\em reg.}) or \bib pairs.} \label{table:ablation_study}
    \centering
    \setlength{\tabcolsep}{3pt}
    \begin{tabular}{c|ccc|c@{\hskip 0.8em}c@{\hskip 0.8em}c@{\hskip 0.8em}c@{\hskip 0.8em}c}
    \toprule
         \multirow{2}{*}{DifS} & \multicolumn{3}{c|}{K selection} & \multicolumn{5}{c}{Number of images annotated} \\
          & im. & reg. & \bib & 50 & 100 & 150 & 200 & 250 \\
         \midrule
          &  &  &  & 56.3 \std{0.4} & 58.0 \std{0.5} & 58.9 \std{0.4} & 60.0 \std{0.3} & 60.5 \std{0.4} \\ 
         \checkmark &  &  &  & 56.5 \std{0.4} & 58.4 \std{0.4} & 59.3 \std{0.7} & 60.2 \std{0.4} & 61.1 \std{0.5} \\ 
         \checkmark & \checkmark &  &  & 57.1 \std{0.4} & 58.3 \std{0.5} & 59.3 \std{0.6} & 59.8 \std{0.4} & 60.3 \std{0.4} \\ 
         \checkmark &  & \checkmark &  & \bf 58.4 \std{0.4} & 60.2 \std{0.4} & 61.5 \std{0.6} & 62.6 \std{0.4} & \bf 63.4 \std{0.3} \\ 
          &  &  & \checkmark & 57.9 \std{0.7} & 60.1 \std{0.4} & 61.2 \std{0.5} & 62.1 \std{0.5} & 62.6 \std{0.4}  \\ 
        \checkmark &  &  & \checkmark & \bf 58.5 \std{0.8} & \bf 60.8 \std{0.5} & \bf 61.9 \std{0.4} & \bf 62.9 \std{0.5} & \bf 63.5 \std{0.4} \\ \bottomrule  
    \end{tabular}
        }
\end{table}

\subsection{Comparison of Active Strategies}
In order to compare \bib to baselines, we conduct 5 active learning cycles with a budget of $B=50$ images (1\% of the training set) per cycle on VOC07 and of $B=160$ images (0.2\% of the training set, 2 fully annotated images per class on average) on COCO. We present results in ~\autoref{fig:active_comparison}. The detailed numbers are provided in the supplemental material. It can be seen that the ranking of the examined baseline methods w.r.t. their detection performance is different on the two datasets. This is explained by the fact that the two datasets have different data statistics. COCO dataset contains many cluttered images, with an average of $7.4$ objects in an image, and VOC07 depicts simpler scenes, with an average of only $2.4$ objects. However, \bib consistently improves over other baselines.

Results on VOC07 (\autoref{fig:active_comparison}a) show that \bib and {\em loss} significantly outperform every method in all cycles. \bib also surpasses {\em loss} except in the first cycle. Entropy and variants of {\em random} perform comparably and slightly better than variants of core-set. Balancing the classes consistently improves the performance of random strategy, albeit with a small margin. Interestingly, \bib reaches the performance of {\em random} at 10\% setting ($\approx500$ images) with only $\approx200$ fully-annotated images. Similarly, it needs fewer than $100$ fully annotated images to attain {\em random}'s performance in the 10-shot ($\approx$ 200 images) setting.

\begin{figure}[t]
\CenterFloatBoxes
\begin{floatrow}
\ffigbox[\FBwidth]{}{
    \includegraphics[width=0.37\textwidth]{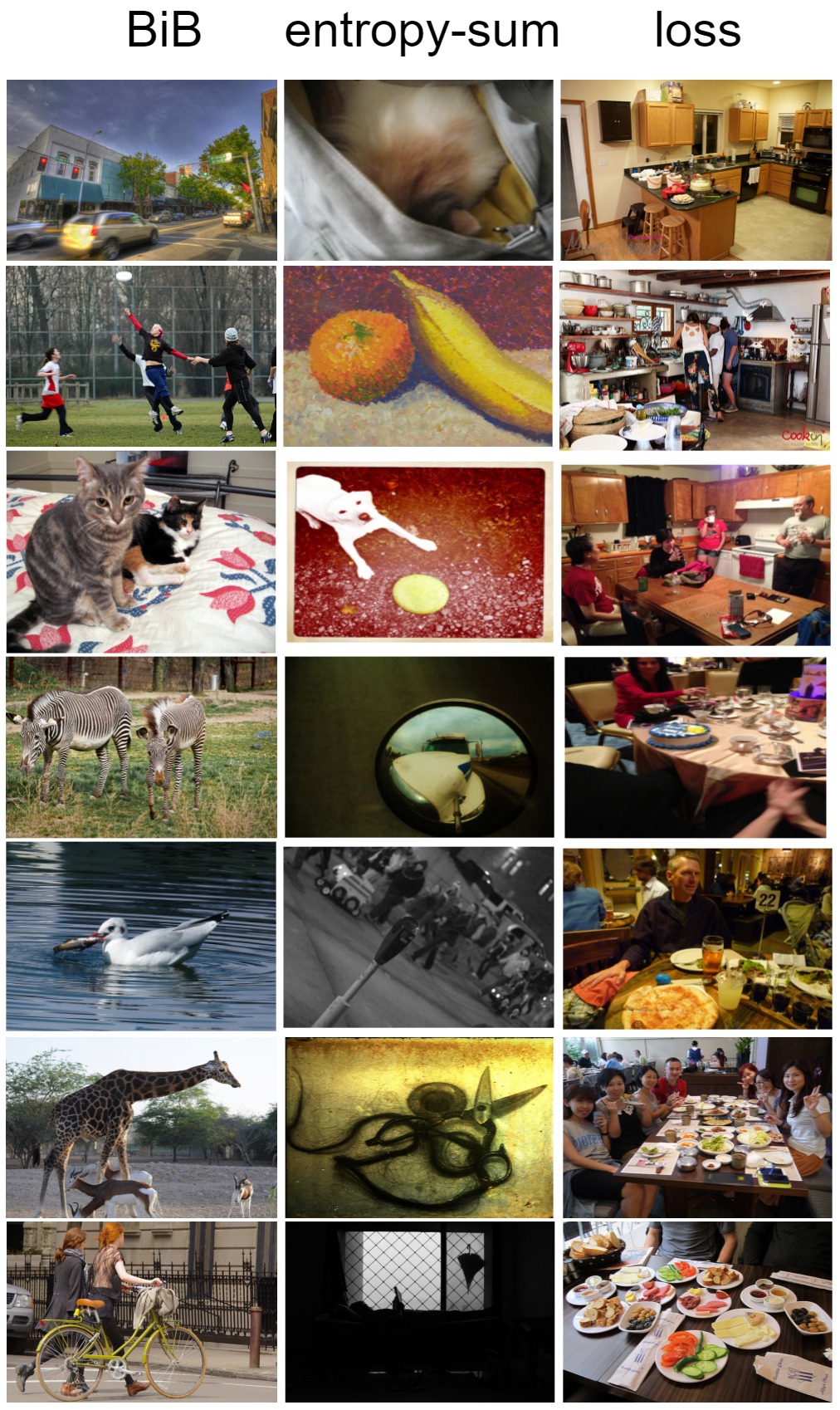}
  \caption{\small 
  Images selected by \bib, {\em entropy-max} and {\em loss} strategies on COCO dataset.}
  \label{fig:selected_images}
}
\capbtabbox{
\resizebox{0.47\textwidth}{!}{
    \begin{tabular}{llccc}
        \toprule
        \multirow{2}{*}{Setting} & \multirow{2}{*}{Method} & \multicolumn{1}{c}{VOC07} & \multicolumn{2}{c}{COCO} \\
         & & \multicolumn{1}{c}{AP50} & AP50 & AP \\
        \midrule
         & & \multicolumn{3}{c}{100\%} \\
        \cmidrule{3-5}
        \multirow{2}{*}{\makecell{Fully\\supervised}} & Fast RCNN~\cite{girshick2015ICCV_fast_rcnn} &\multicolumn{1}{c}{66.9} & 38.6 & 18.9 \\
        & Faster RCNN~\cite{Ren2015NeuRIPS_faster_rcnn} &\multicolumn{1}{c}{69.9} & 41.5 & 21.2 \\
        
        \midrule
         & & \multicolumn{3}{c}{0\%} \\
        \cmidrule{3-5}
        \multirow{6}{*}{WSOD} & WSDDN~\cite{Bilen2016cvpr_WSDDN} & \multicolumn{1}{c}{34.8} & - & - \\
        & OICR~\cite{Tang2017CVPR_oicr_wsod} & \multicolumn{1}{c}{41.2} & - & - \\ 
        & C-MIDN~\cite{Gao2019ICCV_coupled_MIDN_wsod} &  \multicolumn{1}{c}{52.6} & 21.4 & 9.6 \\
        & WSOD2~\cite{zeng2019wsod2} & \multicolumn{1}{c}{53.6} & 22.7 & 10.8 \\
        & MIST-CDB~\cite{ren-cvpr20} & \multicolumn{1}{c}{54.9} & 24.3 & 11.4 \\
        & CASD~\cite{huang2020comprehensive} & \multicolumn{1}{c}{56.8}  & 26.4 & 12.8 \\
        
        \midrule
         & & \multicolumn{3}{c}{10-shot} \\
        \cmidrule{3-5}
        \multirow{4}{*}{\makecell{Weak \& \\few strong}} & BCNet~\cite{Pan2019box_correction} & 57.1 & - & - \\
        & OAM~\cite{Biffi2020ManyshotFL} & 59.7 & 31.2 & 14.9 \\
        & Ours (u-rand) & 60.2 & 32.7 & 16.4 \\
        & Ours (BiB) & 62.9 & 34.1 & 17.2 \\
        
        \bottomrule
    \end{tabular} }
    }{\caption{\small 
    Performance of \bib compared to the state of the art on VOC07 ($B=50$) and COCO ($B=160$) datasets. The \textit{10-shot} setting corresponds to $4$ and $5$ AL cycles resp. on VOC07 and COCO. All of the compared methods use VGG16~\cite{Simonyan15} as the backbone. } 
    \label{table:comparison_to_sota}
    }
\end{floatrow}
\end{figure}

On COCO, \bib again shows consistent improvement over competitors. However, surprisingly, {\em loss} fares much worse than \bib and even {\em random}. To understand these results, we present a representative subset of selected images in \autoref{fig:selected_images}. It appears that images selected by the {\em loss} strategy tend to depict complex scenes. Many of them are indoors scenes with lots of objects (people, foods, furniture, ...). The supervision brought by these images is both redundant (two many images for certain classes) and insufficient (no or too few images for others). This result agrees with those obtained in \cite{Choi_2021_ICCV,liu2021influence} on COCO with the predicted loss method \cite{L4AL19}. On the other hand, variants of entropy strategy tend to select very difficult images that are outliers and not representative of the training dataset. They do not perform well on COCO, especially {\em entropy-sum} which obtains significantly worse results than other strategies. This observation is similar to that of~\cite{MIAOD2021}. Diversity-based methods fare better than uncertainty-based methods, with {\em core-set} and {\em core-set-ent} performing much better than {\em entropy} variants. Among the latter two methods, {\em core-set} performs unsurprisingly better than {\em core-set-ent}, given {\em entropy}'s bad performance. \bib outperforms all other methods. It obtains significantly better results than {\em random}, which other methods fail to do. In addition, \bib attains the same performance as {\em u-random} (see dashed line) with only half as many annotated images, reducing the performance gap (in AP50) between the base weak detector and the fully-supervised Fast RCNN by nearly 70$\%$ with only ten fully annotated images per class on average. It can be seen in \autoref{fig:selected_images} that \bib selects a diverse set of images that reflect the model's confusion on object extent. 

\definecolor{DarkGray}{cmyk}{0,0,0,0.80} 
\definecolor{Plum}{cmyk}{0.50,1.,0,0}
\definecolor{OliveGreen}{cmyk}{0.64,0,0.95,0.40} 
\definecolor{Aquamarine}{cmyk}{0.82,0,0.30,0}
\definecolor{auburn}{rgb}{0.43, 0.21, 0.1}

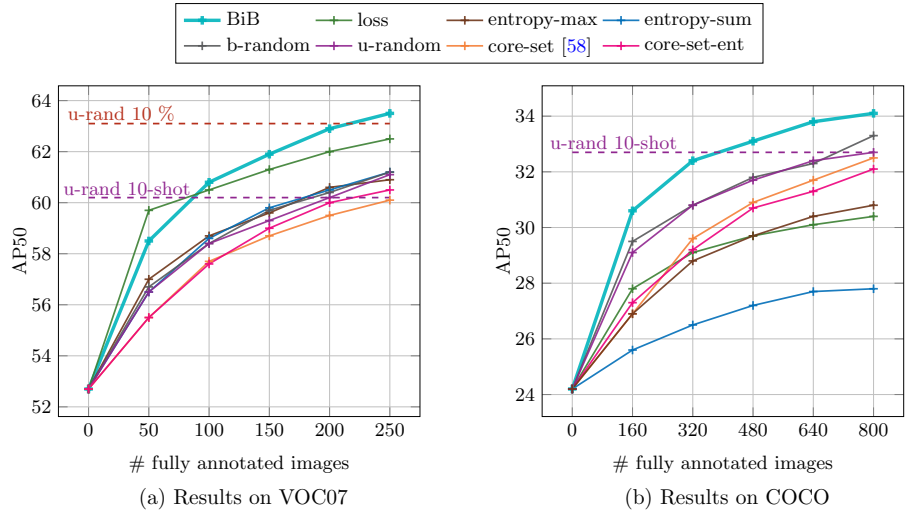
\begin{figure}[t]
    \resizebox{\columnwidth}{!}{
    \begin{tikzpicture} 
        \begin{axis}[
            name=ax1,
            title= {(a) Results on VOC07}, 
            title style={anchor=north,yshift=-190,xshift=2,font=\normalsize},
            height=7cm, width=7.5cm, grid=major,
	        font=\footnotesize,
            xlabel={$\#$ fully annotated images}, 
            ylabel={AP50}, 
            x label style={at={(axis description cs:0.5,-0.095)},anchor=south},
            y label style={at={(axis description cs:0.1,.5)}},
	        xtick={0,50,100,150,200,250},
	        legend cell align={left},
	        legend style={legend columns=4, font=\small, at={(1.95,1.25)},anchor=north east} 
	        ]
            \addplot [opacity=.9,Aquamarine,mark=+,ultra thick] coordinates { (0,52.7) (50,58.5) (100,60.8) (150,61.9) (200,62.9) (250,63.5)}; \addlegendentry{\bib}
            \addplot [opacity=.9,OliveGreen,mark=+,thick] coordinates { (0,52.7) (50,59.7) (100,60.5) (150,61.3) (200,62.0) (250,62.5)}; \addlegendentry{loss}
            
            \addplot [opacity=.9,auburn,mark=+,thick] coordinates { (0,52.7) (50,57.0) (100,58.7) (150,59.6) (200,60.6) (250,60.9)}; \addlegendentry{entropy-max}
            
            \addplot [opacity=.9,NavyBlue,mark=+,thick] coordinates { (0,52.7) (50,56.5) (100,58.6) (150,59.8) (200,60.5) (250,61.2)}; \addlegendentry{entropy-sum}
            \addplot [opacity=.9,DarkGray,mark=+,thick] coordinates { (0,52.7) (50,56.7) (100,58.4) (150,59.7) (200,60.4) (250,61.2)}; \addlegendentry{b-random}
            \addplot [opacity=.9,Plum,mark=+,thick] coordinates { (0,52.7) (50,56.5) (100,58.4) (150,59.3) (200,60.2) (250,61.1)}; \addlegendentry{u-random}
            \addplot [opacity=.9,Orange,mark=+,thick] coordinates { (0,52.7) (50,55.5) (100,57.7) (150,58.7) (200,59.5) (250,60.1)}; \addlegendentry{core-set \cite{sener2018active}}
            \addplot [opacity=.9,RubineRed,mark=+,thick] coordinates { (0,52.7) (50,55.5) (100,57.6) (150,59.0) (200,60.0) (250,60.5)}; \addlegendentry{core-set-ent}
            
            \addplot [color=BrickRed,dashed,thick] coordinates { (0,63.1) (50,63.1) (100,63.1) (150,63.1) (200,63.1) (250,63.1)};
            \node at (axis cs: 27.2,63.52) {\textcolor{BrickRed}{u-rand 10 $\%$}};
            
            \addplot [color=Plum,dashed,thick] coordinates { (0,60.2) (50,60.2) (100,60.2) (150,60.2) (200,60.2) (250,60.2)};
            \node at (axis cs: 32,60.58) {\textcolor{Plum}{u-rand 10-shot}};
        \end{axis}
        \hfill
        \begin{axis}[
            at={(ax1.south east)},
            title= {(b) Results on COCO}, 
            title style={anchor=north,yshift=-190,xshift=2,font=\normalsize},
            xshift=2cm,
            height=7cm, width=7.5cm, grid=major,
	        font=\footnotesize,
            xlabel={$\#$ fully annotated images}, 
            ylabel={AP50}, 
            x label style={at={(axis description cs:0.5,-0.095)},anchor=south},
            y label style={at={(axis description cs:0.1,.5)}},
	        xtick={0,160,320,480,640,800},
	        ]
            
            \addplot [opacity=.9,Aquamarine,mark=+,ultra thick] coordinates { (0,24.2) (160,30.6) (320,32.4) (480,33.1) (640,33.8) (800,34.1)};
            \addplot [opacity=.9,OliveGreen,mark=+,thick] coordinates { (0,24.2) (160,27.8) (320,29.1) (480,29.7) (640,30.1) (800,30.4)}; 
            %
            \addplot [opacity=.9,Orange,mark=+,thick] coordinates { (0,24.2) (160, 26.9) (320, 29.6) (480, 30.9) (640, 31.7) (800, 32.5)};
            
            \addplot [opacity=.9,RubineRed,mark=+,thick] coordinates { (0,24.2) (160, 27.3) (320, 29.2) (480, 30.7) (640, 31.3) (800, 32.1)};
            
            \addplot [opacity=.9,NavyBlue,mark=+,thick] coordinates { (0,24.2) (160,25.6) (320,26.5) (480,27.2) (640,27.7) (800,27.8)}; 
            \addplot [opacity=.9,DarkGray,mark=+,thick] coordinates { (0,24.2) (160,29.5) (320,30.8) (480,31.8) (640,32.3) (800,33.3)}; 
            \addplot [opacity=.9,Plum,mark=+,thick] coordinates { (0,24.2) (160,29.1) (320,30.8) (480,31.7) (640,32.4) (800,32.7)}; 
            
            \addplot [opacity=.9,auburn,mark=+,thick] coordinates { (0,24.2) (160, 26.9) (320, 28.8) (480, 29.7) (640, 30.4) (800, 30.8)};
            
            \addplot [color=Plum,dashed,thick] coordinates { (0,32.7) (160,32.7) (320,32.7) (480,32.7) (640,32.7) (800,32.7)}; 
            \node at (axis cs: 110,33.075) {\textcolor{Plum}{u-rand 10-shot}};
            
        \end{axis}
        \end{tikzpicture}
    \caption{\small 
    Detection performances of different active learning strategies in our framework on VOC07~\cite{pascal-voc-2007} (a) and COCO datasets~\cite{Lin2014cocodataset} (b). We perform $5$ annotation cycles for each strategy with the budget of $B=50$ on VOC07 and $B=160$ on COCO. This corresponds 
    to annotating 1\% and 0.2\% of the training set per cycle respectively for VOC07 and COCO. Dashed lines in \textcolor{Plum}{purple} and \textcolor{BrickRed}{red} highlight results obtained with \emph{10-shot} and \emph{10$\%$} images  selected with \emph{u-random}. Best viewed in color.}
    \label{fig:active_comparison}
}
\end{figure}

\subsection{Comparison to the State of the Art}
We compare the 10-shot performance of our proposed method to the state of the art in \textcolor{red}{Table}~\ref{table:comparison_to_sota}. For \bib, we report the performance of previous experiments (\autoref{fig:active_comparison}) at cycle 4 on VOC07 and cycle 5 on COCO. All compared methods use a Fast R-CNN or Faster R-CNN architecture with a VGG16~\cite{Simonyan15} backbone. Most related to us, OAM~\cite{Biffi2020ManyshotFL} and BCNet~\cite{Pan2019box_correction} also seek to improve the performance of weakly-supervised detectors with a few fully-annotated images. We can see that \bib significantly outperforms them in this setting. In particular, on COCO, we observe from \textcolor{red}{Table}~\ref{table:comparison_to_sota} and \textcolor{red}{Figure}~\ref{fig:active_comparison} that \bib obtains comparable results to 10-shot OAM with only 2 shots (160 images) and significantly better results with 4 shots. Similarly, on VOC07, \bib surpasses the performance of OAM with only a half of the number of fully-annotated images used by the latter. We additionally consider the $10\%$ setting and compare \bib to other baselines on the VOC07 dataset (see Table~\ref{table:per_class_study}). In this setting, a random selection following our method (`Ours (u-rand)') gives an AP50 of 63.1, outperformed by \bib (`Ours (\bib)') which achieves an AP50 of 65.1. In comparison, our main competitors perform worse: OAM (63.3), BCNet (61.8), EHSOD~\cite{Fang_Xu_Liu_Parisot_Li_2020} (55.3) and BAOD~\cite{Pardo2021BAODBO} (50.9).

Compared to WSOD methods, we obtain significantly better results with a small amount of full annotations. \bib enables a greater boost over weakly-supervised detectors than {\em random} and narrows significantly the performance gap between weakly-supervised detectors and fully-supervised detectors. It reduces
the gap between the state of the art weakly-supervised detector CASD~\cite{huang2020comprehensive} and Fast RCNN~\cite{girshick2015ICCV_fast_rcnn} by $5.5$ times with $10\%$ of the training images fully annotated on VOC07 and by $3.5$ times with only $10$ fully annotated images on average per class on COCO. This is arguably a better trade-off between detection performance and data efficiency than both weakly- and fully-supervised detectors. 

\begin{table}[t]
    \LARGE
    \centering
    \resizebox{\textwidth}{!}{%
    \begin{tabular}{l|c|cccccccccccccccccccc|c}
        \toprule
    	method & sup. & aero & bike & bird & boat & bottl & bus & car & cat & chair & cow & table & dog & horse & moto & pers & plant & sheep & sofa & train & tv & mean \\
    	\midrule
    	MIST*& \xmark & 69.0 & 75.6 & 57.4 & 22.5 & 24.8 & 71.5 & 76.1 & 55.9 & 27.6 & 70.3 & 43.9 & 37.5 & 50.8 & \bf 75.9 & 18.5 & 23.9 & 60.8 & 54.7 & 69.3 & 68.1 & 52.7 \\
    	\midrule
        BAOD~\cite{Pardo2021BAODBO} & 10\% & 51.6 & 50.7 & 52.6 & 41.7 & 36.0 & 52.9 & 63.7 & 69.7 & 34.4 & 65.4 & 22.1 & 66.1 & 63.9 & 53.5 & 59.8 & 24.5 & 60.2 & 43.3 & 59.7 & 46.0 & 50.9 \\
     	BCNet~\cite{Pan2019box_correction} & 10\% & 64.7 & 73.1 & 55.2 & 37.0 & 39.1 & \bf 73.3 & 74.0 & 75.4 & 35.9 & 69.8 & 56.3 & \bf 74.7 & 77.6 & 71.6 & 66.9 & 25.4 & 61.0 & 61.4 & \bf 73.8 & 69.3 & 61.8 \\
     	OAM~\cite{Biffi2020ManyshotFL} & 10\% & 65.6 & 73.1 & 59.0 & \bf 49.4 & 42.5 & 72.5 & 78.3 & \bf 76.4 & 35.4 & 72.3 & 57.6 & 73.6 & \bf 80.0 & 72.5 & \bf 71.1 & 28.3 & 64.6 & 55.3 & 71.4 & 66.2 & 63.3 \\
    	Ours (u-r.) & 10\% & \bf 70.5 & 77.2 & 62.3 & 38.5 & 38.5 & 72.3 & \bf 79.4 & 73.6 & 38.6 & 73.8 & 55.7 & 66.5 & 71.4 & 75.3 & 65.5 & 33.8 & 65.4 & 62.7 & 72.3 & 69.7 & 63.1 \\
     	Ours (\bib) & 10\% & 68.9 & \bf 78.1 & \bf 62.7 & 41.4 &\bf 47.8 & 72.4 & \bf 79.2 & 70.3 & \bf 44.9 & \bf 74.7 & \bf 66.2 & 62.2 & 72.1 & \bf 75.6 & 69.8 & \bf 43.1 & \bf 66.2 & \bf 65.0 & 71.4 & \bf 70.7 & \bf 65.1 \\
     	\bottomrule
    \end{tabular}
    }
    \caption{\small Per-class AP50 results on VOC07. \bib yields significant boosts in hard classes such as {\em bottle}, {\em chair}, {\em table} and {\em potted plant}. Results of MIST are the average of three runs using the authors' public code and differ from the numbers in the original paper.}
    \label{table:per_class_study}
\end{table}

\paragbf{Per-class study.} Additionally, 
we present in \textcolor{red}{Table}~\ref{table:per_class_study} the per-class results for different methods on VOC07. It can be seen that variants of our approach (u-random and \bib) consistently boost the 
performance on all classes over MIST~\cite{ren-cvpr20} (except on {\em aeroplane} and {\em motorbike} where they perform slightly worse than MIST). Notably, \bib  yields larger boosts on {\em hard} classes such as {\em table} ($+23$ points \wrt our baseline MIST), {\em chair} ($+17.3$), {\em bottle} ($+23$) and {\em potted plant} ($+19.2$). On those classes, a random selection with our approach is worse than \bib by more than 7 points. 
Overall, \bib obtains the best results on most classes.

\section{Conclusion and Future Work}
We propose a new approach to boost the performance of weakly-supervised detectors using a few fully annotated images selected following an active learning process. We introduce \bib, a new selection method specifically designed to tackle failure modes of weakly-supervised detectors and show a significant improvements over random sampling. Moreover, \bib is effective on both VOC07 and COCO datasets, narrowing significantly the performance between weakly- and fully-supervised object detectors, and outperforming all methods mixing many weak and a few strong annotations in the low annotation regime. 

In this work, we combine weakly-supervised and active learning for reducing human annotation effort for object detectors. There are other types of methods that require no annotation at all, such as unsupervised object discovery \cite{LOST,Vo2020largescale_uod} and self-supervised pre-training \cite{caron2021emerging,chen2020mocov2}, that could help improving different component of our pipeline, e.g., region proposals or the detection architecture. 
Future work will be dedicated to improving our approach by following those directions.

\paragbf{Acknowledgements.} This work was supported in part by the Inria/NYU collaboration, the Louis Vuitton/ENS chair on artificial intelligence and the French government under management of Agence Nationale de la Recherche as part of the ``Investissements d’avenir'' program, reference ANR19-P3IA-0001 (PRAIRIE 3IA Institute). It was performed using HPC resources from GENCI–IDRIS (Grant 2021-AD011013055). Huy V. Vo was supported in part by a Valeo/Prairie CIFRE PhD Fellowship.

\clearpage
%
%

\clearpage
\begin{center}
    \textbf{\large Supplementary materials: Active Learning Strategies for Weakly-Supervised Object Detection}
\end{center}
\setcounter{section}{0}
\setcounter{equation}{4}
\setcounter{figure}{4}
\setcounter{table}{3}
\section{Additional Qualitative Results}
We provide in this section visualizations to get insights into the benefits of our method \bib.

\begin{figure}
\begin{tabular}{lcccc} 
    \centering
    \raisebox{+1.7\normalbaselineskip}[0pt][0pt]{\rotatebox{90}{MIST}} &
    \includegraphics[width=2.8cm,height=2.1cm]{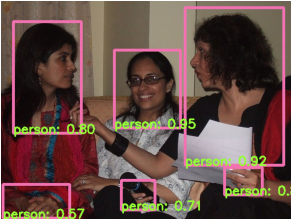}
    \includegraphics[width=2.8cm,height=2.1cm]{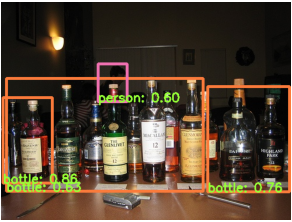}
    \includegraphics[width=2.8cm,height=2.1cm]{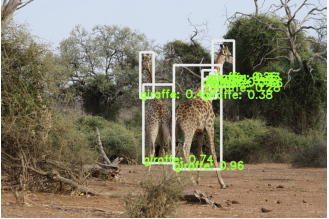}
    \includegraphics[width=2.8cm,height=2.1cm]{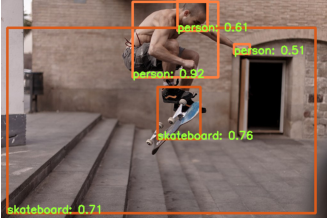} \\
    \raisebox{+.62\normalbaselineskip}[0pt][0pt]{\rotatebox{90}{MIST+BiB}} &
    \includegraphics[width=2.8cm,height=2.1cm]{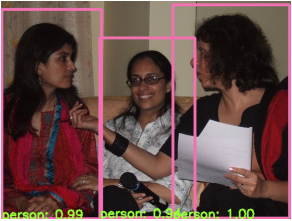}
    \includegraphics[width=2.8cm,height=2.1cm]{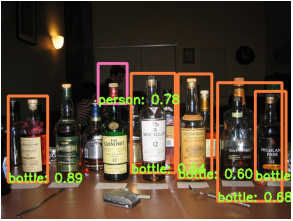}
    \includegraphics[width=2.8cm,height=2.1cm]{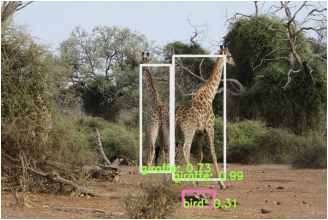}
    \includegraphics[width=2.8cm,height=2.1cm]{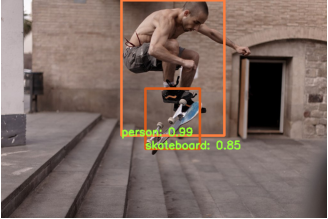}
    \end{tabular}
    \caption{\small Examples of predictions on the VOC07 and COCO test sets, by MIST~\cite{ren-cvpr20} (first row) and \bib after the first cycle (second row). Fine-tuning MIST with images selected by \bib significantly remedies its limitations.}
    \label{fig:preds}
\end{figure}

\subsection{Prediction Examples}
We show in \autoref{fig:preds} predictions obtained with the weakly-supervised detector MIST (top row) and the detector after the first cycle of \bib (bottom row) with $B=50$ on VOC07 and $B=160$ on COCO. We observe that the failures modes of MIST are corrected by our BiB detector: objects and parts are not confused ($3^{rd}$ and $4^{th}$ images), objects are covered ($1^{st}$) and better separated ($2^{nd}$).

\begin{figure}[htb]
    \centering
    \includegraphics[width=4cm,height=3cm]{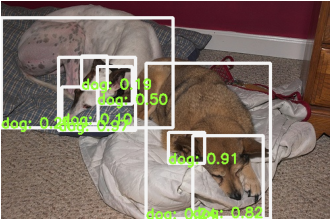}
    \includegraphics[width=4cm,height=3cm]{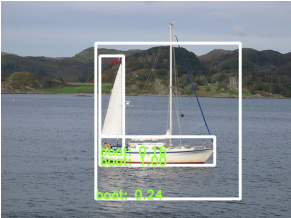}
    \includegraphics[width=4cm,height=3cm]{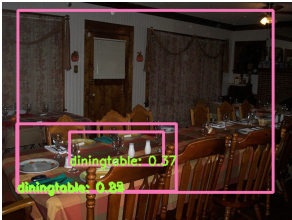}
    
    \includegraphics[width=4cm,height=3cm]{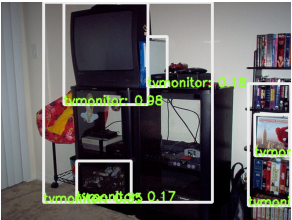}
    \includegraphics[width=4cm,height=3cm]{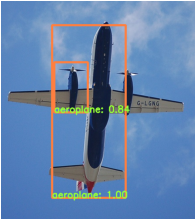}
    \includegraphics[width=4cm,height=3cm]{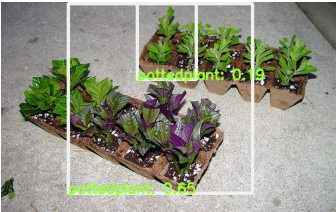}
    
    \includegraphics[width=4cm,height=3cm]{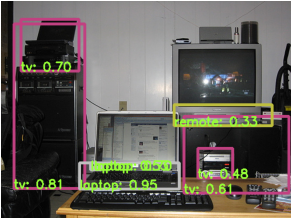}
    \includegraphics[width=4cm,height=3cm]{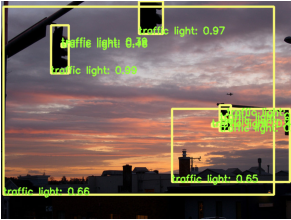}
    \includegraphics[width=4cm,height=3cm]{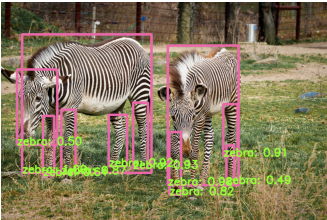}
    
    \includegraphics[width=4cm,height=3cm]{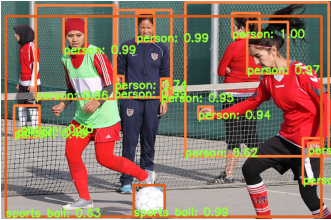}
    \includegraphics[width=4cm,height=3cm]{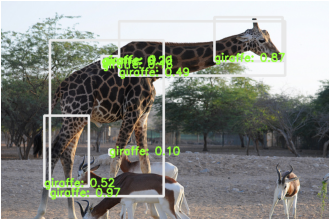}
    \includegraphics[width=4cm,height=3cm]{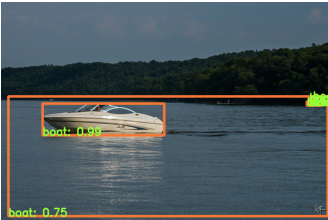}
    
    \caption{\small Examples of \emph{box-in-box} (\bib) pairs on VOC07 (first two rows) and COCO (last two rows) extracted using the MIST~\cite{ren-cvpr20} detector.}
    \label{fig:bib_pairs_sup}
\end{figure}

\subsection{More Visualization of BiB Pairs}
Our selection method relies on the discovery of \emph{box-in-box} patterns.
We provide in \autoref{fig:bib_pairs_sup} more visualization of BiB pairs on images of VOC07 and COCO. 

\begin{table}[t]
    \LARGE
    \centering
    \resizebox{0.7\textwidth}{!}{
    \begin{tabular}{l@{\hskip 0.4em}|@{\hskip 0.4em}c@{\hskip 1.2em}c@{\hskip 1.2em}c@{\hskip 1.2em}c@{\hskip 1.2em}c}
        \toprule
        \multirow{2}{*}{Method} & \multicolumn{5}{c}{Number of fully-annotated images}\\
         & 50 & 100 & 150 & 200 & 250 \\
        \midrule
        u-random & 56.5 \std{0.4} & 58.4 \std{0.4} & 59.3 \std{0.7} & 60.2 \std{0.4} & 61.1 \std{0.5} \\
        b-random & 56.7 \std{0.7} & 58.4 \std{0.7} & 59.7 \std{0.8} & 60.4 \std{0.5} & 61.2 \std{0.4} \\
        core-set & 55.5 \std{0.6} & 57.7 \std{0.6} & 58.7 \std{0.5} & 59.5 \std{0.4} & 60.1 \std{0.2} \\
        core-set-ent & 55.5 \std{0.4} & 57.6 \std{0.4} & 59.0 \std{0.4} & 60.0 \std{0.2} & 60.5 \std{0.2} \\
        entropy-max & 57.0 \std{0.4} & 58.7 \std{0.2} & 59.6 \std{0.4} & 60.6 \std{0.2} & 60.9 \std{0.2} \\
        entropy-sum & 56.5 \std{1.0} & 58.6 \std{0.4} & 59.8 \std{0.3} & 60.5 \std{0.5} & 61.2 \std{0.8} \\
        loss & \bf 59.7 \std{0.2} & 60.5 \std{0.5} & 61.3 \std{0.7} & 62.0 \std{0.5} & 62.5 \std{0.3} \\
        BiB & 58.5 \std{0.8} & \bf 60.8 \std{0.5} & \bf 61.9 \std{0.4} & \bf 62.9 \std{0.5} & \bf 63.5 \std{0.4} \\
        \bottomrule
    \end{tabular}
    }
    \caption{\small Comparison of active learning strategies on VOC07. For each experiment, we conducted $5$ cycles with a budget of $50$ images per cycle. We repeated the experiment six times for each strategy and report the average and standard deviation of their performance.
    \bib yields significantly better performance than the others. {\em loss} performs well in the first cycle but fares worse than \bib in subsequent cycles. Additionally, it performs much worse, even than random, on COCO (see~\autoref{table:active_coco}).}
    \label{table:active_voc}
\end{table}

\begin{table}[t]
    \LARGE
    \centering
    \resizebox{\textwidth}{!}{
    \begin{tabular}{l|@{\hskip 0.4em}ccccc@{\hskip 0.4em}|@{\hskip 0.4em}ccccc}
    \toprule
    \multirow{2}{*}{Method} & \multicolumn{5}{c|@{\hskip 0.4em}}{AP} & \multicolumn{5}{c}{AP50} \\
     & 160 & 320 & 480 & 640 & 800 & 160 & 320 & 480 & 640 & 800 \\
    \midrule
    u-random & 14.1 \std{0.1} & 15.1 \std{0.2} & 15.7 \std{0.2} & 16.1 \std{0.4} & 16.5 \std{0.3} & 29.1 \std{0.4} & 30.8 \std{0.3} & 31.7 \std{0.4} & 32.4 \std{0.4} & 33.0 \std{0.3} \\
    b-random & 14.4 \std{0.4} & 15.2 \std{0.3} & 15.9 \std{0.1} & 16.2 \std{0.2} & 16.8 \std{0.2} & 29.5 \std{0.6} & 30.8 \std{0.4} & 31.8 \std{0.2} & 32.3 \std{0.1} & 33.3 \std{0.2} \\
    entropy-sum & 12.3 \std{0.3} & 12.8 \std{0.2} & 13.3 \std{0.3} & 13.6 \std{0.4} & 13.7 \std{0.3} & 25.6 \std{0.4} & 26.5 \std{0.1} & 27.2 \std{0.2} & 27.7 \std{0.5} & 27.8 \std{0.1} \\
    entropy-max & 12.7 \std{0.2} & 13.9 \std{0.1} & 14.5 \std{0.5} & 14.9 \std{0.3} & 15.2 \std{0.2} & 26.9 \std{0.2} & 28.9 \std{0.1} & 29.7 \std{0.5} & 30.4 \std{0.3} & 30.8 \std{0.3} \\
    loss & 13.5 \std{0.1} & 14.1 \std{0.2} & 14.5 \std{0.2} & 14.7 \std{0.3} & 14.9 \std{0.3} & 27.8 \std{0.1} & 29.1 \std{0.1} & 29.7 \std{0.1} & 30.1 \std{0.3} & 30.4 \std{0.3} \\ 
    core-set & 12.9 \std{0.2} & 14.5 \std{0.3} & 15.3 \std{0.2} & 15.9 \std{0.1} & 16.4 \std{0.3} & 26.9 \std{0.3} & 29.6 \std{0.5} & 30.9 \std{0.2} & 31.7 \std{0.2} & 32.5 \std{0.4} \\
    core-set-ent & 13.1 \std{0.0} & 14.2 \std{0.1} & 15.1 \std{0.2} & 15.5 \std{0.3} & 16.0 \std{0.2} & 27.3 \std{0.2} & 29.2 \std{0.1} & 30.7 \std{0.2} & 31.3 \std{0.4} & 32.1 \std{0.2} \\
    \bib & \bf 14.8 \std{0.3} & \bf 15.9 \std{0.2} & \bf 16.5 \std{0.1} & \bf 16.9 \std{0.2} & \bf 17.2 \std{0.2} & \bf 30.6 \std{0.1} & \bf 32.4 \std{0.3} & \bf 33.1 \std{0.2} & \bf 33.8 \std{0.1} & \bf 34.1 \std{0.1} \\
           
    \bottomrule
    \end{tabular}
    }
    \caption{\small Comparison of active learning strategies on COCO. For each experiment, we conducted $5$ cycles with a budget of $160$ images per cycle. We repeated the experiment three times for each strategy and report the average and standard deviation of their performance. \bib significantly outperforms all other methods.}
    \label{table:active_coco}
\end{table}

\section{Additional Quantitative Results}

\subsection{Detailed Results of Active Learning Strategies}
For experiments with active learning strategies, we have run each strategy six times on VOC07 and three times on COCO and reported the average performance in the main paper. For completeness, we provide in~\autoref{table:active_voc} and~\autoref{table:active_coco} both the average and the standard deviation of the detector's performance in these experiments.

\subsection{Different Variants of {\em loss}}
\label{sec:losses}
MIST~\cite{ren-cvpr20} is trained with a combination of losses coming from different heads. The Multiple Instance Learner produces $\mathcal{L}^{\text{MIL}}$ using the ground-truth class information while each refinement head $k \in \{1,2,3\}$ produces the refinement loss $\mathcal{L}_w^{(k)}$ using pseudo objects generated from the previous head. We have tested each of these losses and the combination of the three refinement losses $\sum_{k=1}^3\mathcal{L}_w^{(k)}$ in our experiments with {\em loss} strategy. We present a summary of the results in~\autoref{table:loss}. For each experiment, we have conducted 5 cycles with a budget of $50$ images per cycle on VOC07. On average, $\mathcal{L}_w^{(3)}$ yields the best results on this dataset and we use it for all experiments with the loss strategy in our submission.
\begin{table}[htb]
    \large
    \centering
    \resizebox{0.8\textwidth}{!}{
    \begin{tabular}{l|@{\hskip 0.4em}c@{\hskip 1.2em}c@{\hskip 1.2em}c@{\hskip 1.2em}c@{\hskip 1.2em}c@{\hskip 0.4em}}
        \toprule
         & \multicolumn{5}{c}{Number of fully-annotated images}\\
        AL method & 50 & 100 & 150 & 200 & 250 \\
        \midrule
        $\mathcal{L}^{\text{MIL}}$ & 57.1 \std{0.3} & 57.9 \std{0.2} & 58.4 \std{0.5} & 59.4 \std{0.2} & 60.0 \std{0.3} \\
        $\mathcal{L}_w^{(1)}$ & 58.2 \std{0.4} & 58.5 \std{0.4} & 59.6 \std{0.7} & 60.3 \std{0.8} & 61.1 \std{0.5} \\
        $\mathcal{L}_w^{(2)}$ & 59.4 \std{0.3} & \bf 60.7 \std{0.2} & \bf 61.4 \std{0.3} & 61.8 \std{0.3} & 62.4 \std{0.1} \\
        $\mathcal{L}_w^{(3)}$ & 59.7 \std{0.2} & 60.5 \std{0.5} & 61.3 \std{0.7} & \bf 62.0 \std{0.5} & \bf 62.5 \std{0.3} \\
        $\sum_{k=1,2,3} \mathcal{L}_w^{(k)}$ 
        & \bf 59.9 \std{0.4} & 60.6 \std{0.5} & 60.9 \std{0.5} & 61.6 \std{0.3} & 62.2 \std{0.6} \\
        \bottomrule
    \end{tabular}
    }
    \caption{\small Performance of the loss strategy with different choices of the detector's loss on VOC07. For each experiment, we perform $5$ cycles with a budget of $50$ images per cycle. We have repeated the experiment six times for each strategy and report the average and standard deviation of their performance.} 
    \label{table:loss}
\end{table} 

\begin{table}[htb]
\centering
\small
\setlength{\tabcolsep}{5pt}
\begin{tabular}{c|ccc|ccccc}
\toprule
     \multirow{2}{*}{DifS} & \multicolumn{3}{c|}{K selection} &  \multicolumn{5}{c}{AP50} \\
      & im. & reg. & \bib 
      & 160 & 320 & 480 & 640 & 800 \\
     \midrule
      &  &  &  & 29.0 & 30.6 & 31.4 & 32.3 & 32.8 \\ 
     \checkmark &  &  &  & 29.1 & 30.8 & 31.7 & 32.4 & 33.0  \\ 
     \checkmark & \checkmark &  &  & 29.2 & 30.7  & 31.6  & 32.3  & 32.9  \\ 
     \checkmark &  & \checkmark &  & 30.5 & 31.6  & 32.6 & 33.5  & 34.1 \\ 
      &  &  & \checkmark & 30.7 & 32.3 & 33.2 & 33.7 & 34.2  \\ 
    \checkmark &  &  & \checkmark & 30.6 & 32.4 & 33.1 & 33.8  & 34.1 \\ \bottomrule  
\end{tabular}
\caption{\small \textbf{Ablation study on COCO}. Results in AP50 on COCO with $5$ cycles and a budget $B = 160$. We provide averages and standard deviation results over several runs. \emph{DifS} stands for the difficulty-aware region sampling module. Images are selected by applying k-means++ init. ({\em K selection}) on image-level features ({\em im.}), confident predictions' features ({\em reg.}) or \bib pairs.}
\label{table:ablation_coco}
\end{table}

\subsection{Ablation study on COCO.} 
We have provided an ablation study on different components of \bib on VOC07 dataset in the main paper. For completeness, we report in \textcolor{red}{Table~\ref{table:ablation_coco} } the averaged AP50 scores (over 3 repetitions) 
of the ablation study on COCO. The results are similar to those obtained on VOC, except for the difficulty-aware sampling, which helps with the u-random strategy but not always with \bib.

\subsection{Are diverse samples important?} 
We propose in \bib to find diverse images on which the weakly-supervised detector fails. We investigate the importance of sample diversity in \bib by comparing it to two variants. In the first variant, we randomly select images containing \bib pairs (`U(BiB)'), and in the second variant, we use a mix, with half selected with \bib and the other half with randomly uniform sampling (`U+\bib'), to be fully annotated. 
We show the results in \textcolor{red}{Table}~\ref{table:bib_and_random}.
The fact that U(\bib) is worse than \bib and U+\bib outperforms U(\bib) in general shows that diversity sampling is important once \bib patterns have been discovered.

\begin{table}[htb]
\small 
\centering
\begin{tabular}{l@{\hskip 0.8em}c@{\hskip 0.8em}ccccc}
    \toprule
   \multirow{2}{*}{Method} &  \multirow{2}{*}{Dataset} & \multicolumn{5}{c}{AL cycles} \\
    &  & 1 & 2 & 3 & 4 & 5 \\
   \midrule 
  u-rand. & \multirow{4}{*}{VOC} & 56.5 & 58.4 & 59.3 & 60.2 & 61.1 \\
  U(\bib) & & 57.6 & 59.2 & 60.1 & 61.2 & 61.8 \\
   U+\bib & & 57.9 & 59.4 & 60.7 & 61.6 & 62.4 \\
  \bib & & \bf 58.5 & \bf 60.8 & \bf 61.9 & \bf 62.9 & \bf 63.5  \\
   \midrule
   u-rand & \multirow{4}{*}{COCO} & 29.1 & 30.8 & 31.7 & 32.4 & 33.0 \\
   U(\bib) & & 30.0 & 31.4 & 32.3 & 33.1 & 33.5 \\
   U+\bib & & 29.7 & 31.4 & 32.4 & 33.2 & 33.7 \\
   \bib & & \bf 30.6 & \bf 32.4 & \bf 33.1 & \bf 33.8 & \bf 34.1 \\
   \bottomrule
\end{tabular}
\caption{\small A comparison between \bib, u-rand and two other variants that combine them. \bib outperforms the variants, showing that diversity sampling is important to the effectiveness of \bib.}
\label{table:bib_and_random}
\end{table}

\subsection{Verification of \bib pairs}
We propose in our paper the use of \bib pairs as an indicator of a detector's confusion on images. With its design, we argue that at least one box in the pair is likely a wrong prediction. We verify this assumption on MIST's predictions on VOC07 and COCO. Among $8{,}758$ \bib pairs on VOC, there are $8{,}633$ pairs (98.6\%) with at least one wrong prediction while 99.6\% of the $854{,}004$ \bib pairs have at least one wrong box on COCO.

\subsection{Number of \bib examples reduced with active learning cycles.} 
We use \bib pairs as an indicator of the model's confusion on images. Intuitively, as the model becomes more accurate with more active learning cycles, fewer \bib pairs will be found. We computed the number of \bib pairs during active learning cycles on VOC07 and COCO datasets to verify this assumption. As expected, our results show that it decreases with iterations. On VOC, it drops from 8801 in cycle 1 to 5170 in cycle 5 with budget $B=50$. On COCO, it decreases from 854k in cycle 1 to 152k in cycle 5 with budget $B=160$.

\subsection{Influence of Hyper-Parameters}
We use two intuitive hyper-parameters in \bib design: the area ratio $\mu$ between two boxes in a \bib pair and the
ratio $\delta$ of the overlap over the smallest box. By design, the latter should be close to $1$ so that the small box is ``contained'' in the large box and it is set to $0.8$ in our experiments. 
For the former, we test \bib on VOC07 when its value varies in $\{2,3,4\}$ and report results in \autoref{table:m_value}.
It can be seen that the performance is relatively insensitive to $\mu$. We use $\mu=3$ in our experiments.

\begin{table}[htb]
    \large
    \centering
    \resizebox{0.7\textwidth}{!}{
    \begin{tabular}{c@{\hskip 0.4em}|@{\hskip 0.4em}c@{\hskip 1.2em}c@{\hskip 1.2em}c@{\hskip 1.2em}c@{\hskip 1.2em}c@{\hskip 0.4em}}
        \toprule
         & \multicolumn{5}{c}{Number of fully-annotated images}\\
        $\mu$ & 50 & 100 & 150 & 200 & 250 \\
        \midrule
        $\mu=2$ & 58.5 \std{0.5} & 60.4 \std{0.3} & 61.6 \std{0.4} & 62.4 \std{0.3} & 63.1 \std{0.2} \\
        $\mu=3$ & 58.5 \std{0.8} & 60.8 \std{0.5} & 61.9 \std{0.4} & 62.9 \std{0.5} & 63.5 \std{0.4} \\
        $\mu=4$ & 58.3 \std{0.5} & 60.6 \std{0.3} & 61.7 \std{0.3} & 62.5 \std{0.4} & 63.3 \std{0.2} \\
        \bottomrule
    \end{tabular}
    }
    \caption{\small Performance of \bib on VOC07 with different values of the area ratio $\mu$ in \bib design. We conducted $5$ cycles with a budget of $50$ images per cycle, repeated the experiment six times for each value of $\mu$ and report the average and standard deviation of their performance.} 
    \label{table:m_value}
\end{table} 

\section{More Details}
\subsection{MIST Architecture}
We use MIST~\cite{ren-cvpr20} as our base weakly-supervised object detector and briefly describe it in the main submission. 
MIST follows OICR~\cite{Tang2017CVPR_oicr_wsod} and consists of a Multiple Instance Learner (MIL) trained to produce coarse detections which are then refined with several refinement heads using automatically-generated pseudo-boxes.
We have given details about the refinement heads in the main paper and provide here a description of the MIL head as well as the procedure to generate the pseudo-boxes.
We consider an image $\mathbf{I}$, its class labels $\mathbf{q} \in \{0,1\}^{C}$ and the set of pre-computed region proposals $\mathcal{R}=\{r_1,r_2,\dots,r_R\}$. Please note that we drop here the image index in order to ease understanding.

\subsubsection{Multiple Instance Learner.}
MIL receives $\mathbf{I}$ and $\mathcal{R}$ as input and yields a class probability vector $\phi \in \mathbb{R}^{C}$. It is trained to classify the image with the Binary Cross Entropy (BCE) loss $\mathcal{L}_{\text{MIL}}$ on $\phi$:
\begin{equation}
    \mathcal{L}_{\text{MIL}} = -\dfrac{1}{C}\sum_{c=1}^C q(c)\log(\phi(c)) + (1-q(c))\log(1-\phi(c)).
\end{equation}
In MIST, class probabilities $\phi$ are obtained by aggregating scores in a region score matrix $\textbf{s} \in \mathbb{R}^{R \times C}$ with $c \in \{1,..,C\}$:
\begin{equation}
    \phi(c) = \sum_{i=1}^R \textbf{s}(i,c),
\end{equation}
where $\textbf{s} = \textbf{s}_c \odot \textbf{s}_d$  
is the point-wise product of a
classification score matrix $\textbf{s}_c \in \mathbb{R}^{R \times C}$ and 
a detection score matrix $\textbf{s}_d \in \mathbb{R}^{R \times C}$. 
Matrices $\textbf{s}_c$ and $\textbf{s}_d$ are built by concatenating projected regions features extracted with the backbone network for each of the regions in $\mathcal{R}$. Matrix $\textbf{s}_c$ is normalized row-wise with the softmax operation and models the class probabilities of the region proposals. Matrix $\textbf{s}_d$, which is normalized column-wise, represents the relative objectness of the proposals with respect to the corresponding classes. Given those interpretations, $\textbf{s}(i,c)$ expresses the likelihood that region $i$ is an object of class $c$.

\subsubsection{Pseudo-boxes generation.}
MIST~\cite{ren-cvpr20} introduces a heuristic to generate the pseudo-boxes $\mathbf{D}^{(k-1)}$  that are used to train the refinement heads $k$. Such boxes are generated either from the region score matrix $\textbf{s}$ of the MIL (giving $\mathbf{D}^{(0)}$) or the region classification score matrices $\textbf{s}^{(k)}$ ($k=1,2,3$) of the refinement heads (giving $\mathbf{D}^{(k)}$). 
In particular, for each ground-truth class $c$ in image $\mathbf{I}$, the corresponding column scores $[\textbf{s}(1,c),\dots,\textbf{s}(R,c)]$
in $\textbf{s}$ (or $\textbf{s}^{(k)}$) are sorted in descending order. Then, given the top-15\% region proposals with the highest scores, we select all boxes that do not have an IoU$\geq 0.3$ with a higher-ranked region.
Selected boxes for all classes are aggregated to construct the final set of pseudo-boxes.

\subsection{Active Learning Strategies}
\label{sec:al}
We compare in the main text our proposed \bib to different active learning strategies. We detail here all considered methods. As described in the \textcolor{red}{Algorithm~$1$} of the submission, a set of images $A^t$ of $B$ images is selected at each cycle $t$. The selection is performed with an active learning method within the set of images $W^{t-1}$, possibly using the detector $M^{t-1}$ trained at the end of the previous cycle and the set of selected images $S^{t-1}$.

\paragraph{Random.} We implement two variants of a random sampling: {\em u-random} and {\em b-random}.
In {\em u-random}, $B$ images are selected uniformly at random from $W^{t-1}$; {\em b-random} seeks to have a balance sampling among the classes. 
Images are iteratively selected until the budget $B$ is reached. At each iteration, an image containing at least an object of the class that is the least represented\footnote{In case of draw, a class is randomly selected.} in $S^{t-1} \cup A^t$ is randomly chosen and added to $A^t$.

\paragraph{Diversity-based strategies.} The core-set~\cite{sener2018active} approach attempts to select a representative subset of a dataset.
We employ the greedy version of {\em core-set} in our experiments. In particular, at cycle $t$, let $\psi_{t-1}(\textbf{I}_i)$ be the features of image $\textbf{I}_i$ extracted from detector $M^{t-1}$, {\em core-set} iteratively selects the image $i^*$ to be added in $A^t$
by solving the optimization problem: 
\begin{equation}
	i^* = \underset{i \in W^{t-1}\setminus A^t}{\text{argmax}}\,\underset{j \in S \cup A^t}{\text{min}}
		\|\psi_{t-1}(\mathbf{I}_i) - \psi_{t-1}(\mathbf{I}_j)\|.
\label{eq:coreset}
\end{equation}
In the first cycle, the very first image is randomly selected. 

\paragraph{Selection using model uncertainty.} The concept of \emph{informativeness} has been widely exploited in the literature \cite{MIAOD2021,L4AL19,Brust2019,Choi_2021_ICCV}. 
For a classification task, the uncertainty can be computed by measuring the \emph{entropy} over the class predictions of an image. 
Here, we first compute the entropy over the class predictions of each predicted box in an image,
and then the box entropy-scores of an image are aggregated using the
{\em sum} and {\em max} pooling,
resulting in two strategies, {\em entropy-sum} and {\em entropy-max}. Concretely, let $p_{i,j} \in \mathbb{R}^{C+1}$ be the predicted class scores of the predicted box $j$ for image $\textbf{I}_i$ given by $M^{t-1}$, and $\textbf{D}_i$ be the set of all predictions in $\textbf{I}_i$, we compute the uncertainty score $u_i$ of image $\textbf{I}_i$ as 
\begin{equation}
    \max_{1\leq j \leq |\textbf{D}_i| } \sum_{c=1}^{C+1} - p_{i,j}^c \log(p_{i,j}^c)  
\end{equation}
for {\em entropy-max} and 
\begin{equation}
    \sum_{1\leq j \leq |\textbf{D}_i| } \sum_{c=1}^{C+1} - p_{i,j}^c \log(p_{i,j}^c)   
\end{equation}
for {\em entropy-sum}.
Then, the $B$ images with the highest scores in $\textbf{u}$ are selected.

\paragraph{Combining diversity and uncertainty.} Following \cite{haussmann2020scalable}, we consider a selection strategy function that incorporates the uncertainty information into \emph{core-set} by multiplying the distances between image features 
with the uncertainty score $\textbf{u}$ defined above. 
Specifically we combine \emph{core-set} and \emph{entropy-max}, in a new active learning method {\em core-set-ent} which iteratively selects an image $i^*$ following: 
\begin{equation}
	i^* = \underset{i \in W^{t-1}\setminus A^t}{\text{argmax}}\,\underset{j \in S \cup A^t}{\text{min}} u_i \times
		\|\psi_{t-1}(\mathbf{I}_i) - \psi_{t-1}(\mathbf{I}_j)\|.
\label{eq:coreset-entr}
\end{equation}

\paragraph{Selection using losses.} In \cite{L4AL19}, the authors propose to learn -- through an auxiliary module -- an object detection loss predictor which later allows choosing samples that produce the highest losses. Conveniently, the refinement heads of MIST produce refinement losses ($\mathcal{L}_w^{(k)}$ with $k \in \{1,2,3\}$) that are \emph{detection} losses computed using pseudo-boxes. We therefore propose the active learning method \emph{loss} which 
selects the $B$ images with the highest loss $\mathcal{L}_w^{(3)}$. We have discussed in \textcolor{red}{Section}~\ref{sec:losses} results obtained when considering different losses of MIST.

\end{document}